\documentclass[letterpaper, 10 pt, conference]{ieeeconf}  

\IEEEoverridecommandlockouts                              

\usepackage{amsmath}
\usepackage{amssymb}
\usepackage{mathtools}
\usepackage{graphicx}
\usepackage{float}
\usepackage{array}
\usepackage{tikz}
\usepackage{listings}
\usepackage{hyperref}
\usepackage{graphicx}
\usepackage{cite}
\usepackage{dsfont}
\usepackage{subfig}
\usepackage{mathtools,etoolbox}
\usepackage[ruled, linesnumbered]{algorithm2e}
\usepackage[none]{hyphenat}
\usepackage[utf8]{inputenc}
\usepackage[english]{babel}

\usepackage{amsthm}
\usepackage{xfrac}
\usepackage{nicefrac}
\usepackage{booktabs}
\usepackage{flushend}

\hypersetup{colorlinks=true,urlcolor=blue,citecolor=red}

\title{\LARGE \bf SalientDSO: Bringing Attention to Direct Sparse Odometry}
\author{Huai-Jen Liang, Nitin J. Sanket, Cornelia Ferm{\"u}ller, Yiannis Aloimonos 
\thanks{All authors are associated with University of Maryland, College Park. Emails: \{{\tt\footnotesize hjliang@terpmail, nitin@umiacs, fer@umiacs, yiannis@umiacs}\} {\tt \footnotesize @.umd.edu}}} 

\begin{document}
\makeatletter
\g@addto@macro\@maketitle{
\begin{figure}[H]
   \setlength{\linewidth}{\textwidth}
   \setlength{\hsize}{\textwidth}
    \centering
    \includegraphics[width=\textwidth]{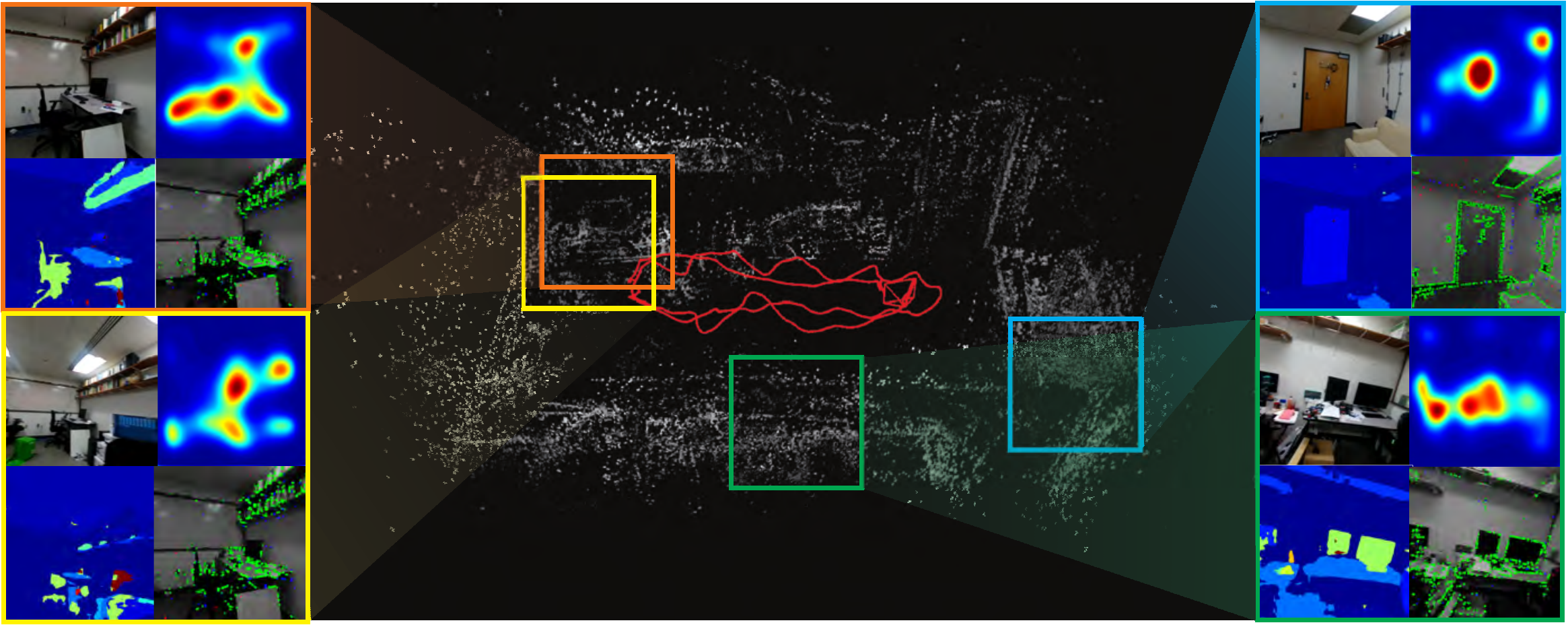}
    \caption{Sample point-cloud output of SalientDSO which does not have loop closure or global bundle adjustment. The insets show the corresponding image, saliency, scene parsing outputs and active features. Observe that features from non-informative regions are almost removed approaching object centric odometry.}
    \label{fig:Overview}
    \end{figure}
}
\maketitle
\thispagestyle{empty}
\pagestyle{empty}


\setcounter{figure}{1}

\begin{abstract}
Although cluttered indoor scenes have a lot of useful high-level semantic information which can be used for mapping and localization, most Visual Odometry (VO) algorithms rely on the usage of geometric features such as points, lines and planes. Lately, driven by this idea, the joint optimization of semantic labels and obtaining odometry has gained popularity in the robotics community. The joint optimization is good for accurate results but is generally very slow. At the same time, in the vision community, direct and sparse approaches for VO have stricken the right balance between speed and accuracy.\\
\indent We merge the successes of these two communities and present a way to incorporate semantic information in the form of visual saliency to Direct Sparse Odometry -- a highly successful direct sparse VO algorithm. We also present a framework to filter the visual saliency based on scene parsing. Our framework,  \textit{SalientDSO}, relies on the widely successful deep learning based approaches for visual saliency and scene parsing which drives the feature selection for obtaining highly-accurate and robust VO even in the presence of as few as 40 point features per frame. We provide extensive quantitative evaluation of SalientDSO on the ICL-NUIM and TUM monoVO datasets and show that we outperform DSO and ORB-SLAM -- two very popular state-of-the-art approaches in the literature. We also collect and publicly release a CVL-UMD dataset which contains two indoor cluttered sequences on which we show qualitative evaluations. To our knowledge this is the first paper to use visual saliency and scene parsing to drive the feature selection in direct VO.

\end{abstract}


\section*{Supplementary Material}
The accompanying video and dataset is available at
\url{prg.cs.umd.edu/SalientDSO.html}.

\section{Introduction and Philosophy}
Simultaneous Localization and Mapping (SLAM) and Visual Odometry (VO) algorithms have taken center stage in the recent years due to their wide-spread usage. They play a prominent part in the perception and planning pipelines of self-driving cars, autonomous quadrotors,  augmented and virtual reality. The never ending quest to come up with realtime solutions for these methods whilst being as accurate as their offline counterparts has led to alternative problem formulations in terms of constraints and optimization methods \cite{monoslam, PTAM, DTAM, LSDSLAM}. 

Not so long ago, the field was dominated by indirect methods \cite{monoslam, PTAM, ORBSLAM, stuhmer2010real} which rely on feature matching and foundations of multi-view geometry coupled with windowed optimization to build a map of the scene and obtain accurate poses. These approaches are based on the low-level geometric features and do not work very well with environments with repeating structures and texture-less surfaces. Some works have improved upon the previous approaches in-terms of speed and accuracy by incorporating prior knowledge such as the dynamics of the system and/or data from more sensors such as inertial measurement units \cite{ROVIO}, time-of-flight sensors \cite{RGBDSLAM} etc. However, minimalism is a trend forward, i.e., trying to achieve the same tasks with a minimal number of sensors. In the scope of this paper, we focus on a monocular VO solution. The current state-of-the-art in monocular approaches which have the best compromise of speed and accuracy are direct sparse approaches such as Direct Sparse Odometry (DSO)\cite{DSO}.

However, object centric SLAM approaches are more robust by nature due to the high level semantics used in the formulation. Lately, joint optimization of 3D poses, stucture and labelled object locations has improved the state-of-the-art significantly. These frameworks rely on the widely successful deep learning based object recognition engine and pose graph optimization frameworks, combining both low-level geometric features and the high-level semantics.

However, humans perform the task of mapping very differently. The human visual system interprets the scene for various tasks like recognition, segmentation, tracking and navigation by making a series of fixations \cite{YiannisFixation}. This is called the Active approach \cite{ActiveVision, SukhtameActive, BajcsyActive}, whilst the traditional approach is called the Passive approach (See Table \ref{tab:ActiveVsPassive}). These fixations lie in the proto-segmentation of the salient objects/locations in the scene. The word proto-segmentation refers to the fact that a segmentation around the fixation point may lead to partial/complete segmentation of an object, which depends on the scenario. Solving the problem of recognition and tracking along with segmentation is like a chicken-egg problem. One would need a good segmentation for recognition and tracking and vice-versa. An Expectation-Maximization (EM) type of scheme, where one would jointly/alternatively optimize for the segmentation and recognition/tracking has gained popularity in literature lately, due to the advancement of fast and accurate optimization frameworks.


\begin{table}[t!]
\centering
\caption{Active vs Passive approach for computer vision tasks.}
\resizebox{\columnwidth}{!}{
\label{tab:ActiveVsPassive}
\begin{tabular}{p{0.2\columnwidth}p{0.4\columnwidth}p{0.4\columnwidth}}
\toprule
Task & Passive approach & Active approach \\
 \hline \\
 Segmentation & Graph cut or super-pixel based methods. & Fixation based region segmentation and recognition in a feedback loop. \\
Recognition & Sliding window of filter banks with a classification algorithm for final prediction. &  Saliency/fixation based segmentation/clustering followed by selection of attributes and sliding window of filters with a simple classification algorithm.\\
Tracking and Failure recovery & Making an online dictionary for robustness against changes and use detection for failure recovery. & Tightly couple saliency into the tracking filter to reduce search space and use salient regions for failure recovery. By doing so, we introduce high level semantics into the low level processes (feedback).\\
Navigation and Mapping & Map based on features based on image gradients.  & Map only using salient region features or objects obtained using fixation based segmentation. Take advantage of the semantic relationships between differently labeled regions.\\
 \bottomrule
\end{tabular}}
\end{table}

Very recently, this philosophy of fixation and attention has started to gain popularity in the robot navigation community \cite{SemanticSLAM, KostasSemanticSLAM, an2017semantic, alexis}. This is based on the fact that humans perform the task of mapping very differently from how it has been done in the robotics literature. They build ``sematic/toplogical'' maps to traverse the scene. This paper combines the concepts used by humans and robotics literature to present a framework of indoor visual odometry in which the features are selected based on a visual saliency map that is obtained by human eye tracking data. This work aims to mimic the qualitative human vision in the framework of direct VO. The key contributions of this paper are:

\begin{itemize}
    \item We present a framework of indoor visual odometry in which the features are selected based on a visual saliency map (Sample output is shown in Fig. \ref{fig:Overview}).
    \item We present a method to filter saliency map based on scene parsing.
    \item We provide experimental results on various simulated and real indoor environments to demonstrate the improved performance of the proposed approach with comparisons to the state-of-the-art.
\end{itemize}
The rest of the paper is organized as follows: Sec. \ref{sec:SalientDSO} presents the different parts of the proposed SalientDSO framework along with the preliminaries required. Sec. \ref{sec:PointSel} describes the visual saliency and scene parsing driven point selection algorithm used in SalientDSO. Detailed experiments along with quantitative and qualitative results are given in Sec. \ref{sec:results}. We finally conclude the paper in Sec. \ref{sec:Conc} with parting thoughts on future work.

\section{SalientDSO Framework}
\label{sec:SalientDSO}
\begin{figure}[t!]
    \centering
    \includegraphics[width=\columnwidth]{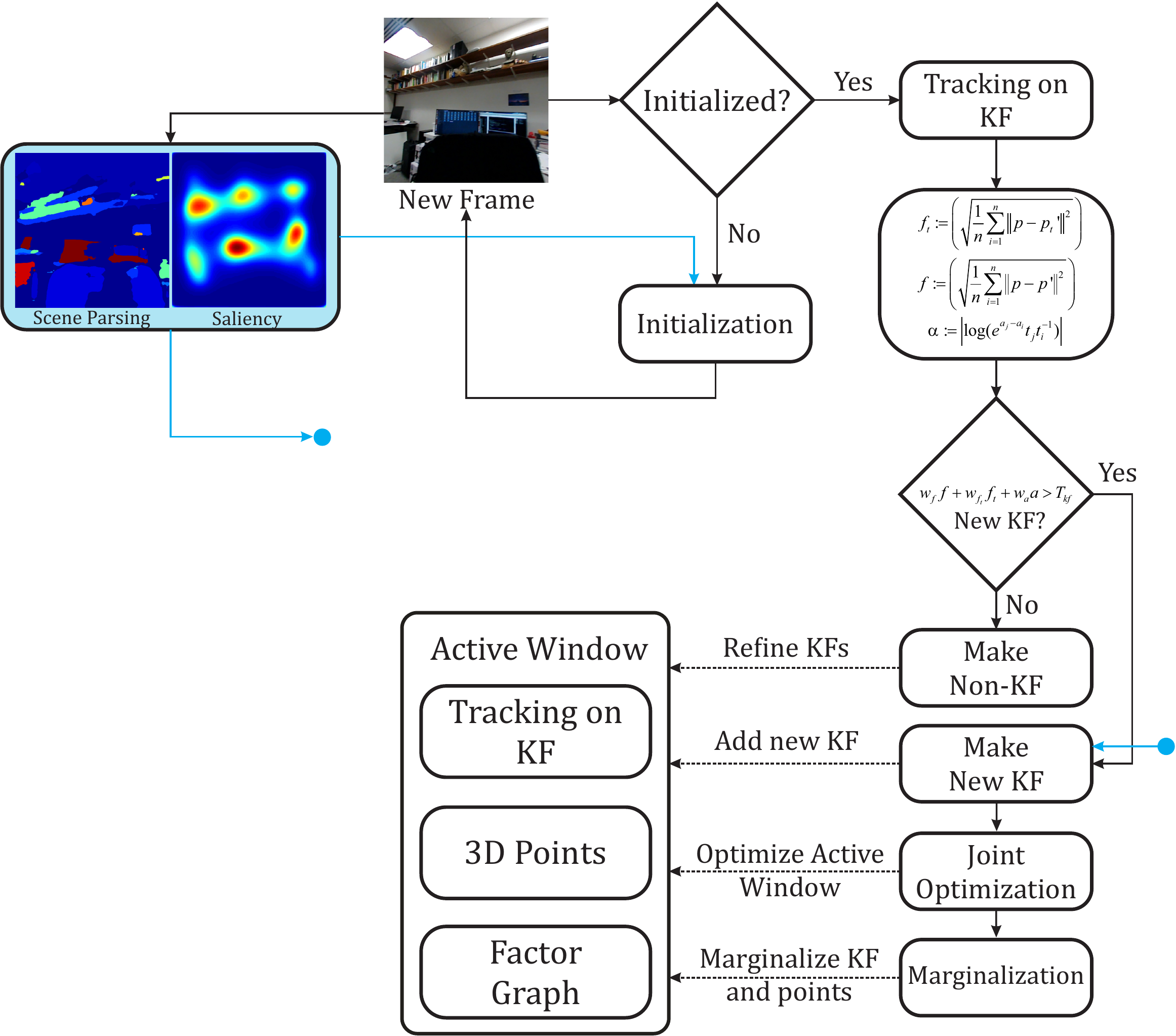}
    \caption{Algorithmic overview of SalientDSO, blue parts show our contributions.}
    \label{fig:FlowChart}
\end{figure}

SalientDSO's framework is composed of a pre-processing step and a VO backbone. The VO backbone is responsible for initializing and tracking camera pose and optimizing all model parameters. The pre-processing step involves the saliency prediction and scene parsing using  deep Convolutional Neural Networks (CNNs) and later using these outputs to select features/points. Fig. \ref{fig:FlowChart} shows the algorithmic overview of SalientDSO, where blue parts of the figure show our contributions (which constitute the pre-processing step). Each component of SalientDSO is discussed in detail next.\\

\subsection{Visual Odometry Backbone}
We adopt DSO \cite{DSO} as the backbone VO in SalientDSO. In brief, DSO\cite{DSO} proposed a direct sparse model to jointly optimize all parameters (camera intrinsics, camera extrinsics, and inverse-depth values for feature points) and perform windowed bundle adjustment. It contains a front end for frames/points selection and initialization, and a back end for optimization. In the proposed framework, points selection is replaced with our proposed method in Sec.\ref{sec:points}. The front-end and back-end are detailed next.\\

\subsubsection{Front-end}
The front-end part of algorithm handles the following:

\textbf{Initial Frame Tracking:} A new frame is tracked with respect to the latest KeyFrame (KF) by using the conventional two-frame direct image alignment, a multi-scale image pyramid and a constant motion model. If tracking fails, SalientDSO attempts to recover the motion by trying 27 different small rotations.

\textbf{Keyframe Creation:} Similar to ORB-SLAM\cite{ORBSLAM}, SalientDSO initially takes many keyframes (around 5-10 keyframes per second), and then sparsifies them by early marginalization of the redundant KFs. SalientDSO uses the following three rules to decide if a new KF is needed: mean square optical flow $\left({f_t}: = \left( {\tfrac{1}{n}\sum_{i = 1}^n {{{\left\| {p - {p_t}'} \right\|}^2}} } \right)^{0.5}\right)$, mean flow without rotation $\left( f: = \left( {\tfrac{1}{n}\sum_{i = 1}^n {{{\left\| {p - p'} \right\|}^2}} } \right)^{0.5}\right)$ and relative brightness factor $\left( \alpha : = \left| {\log \left( {{e^{{a_j} - {a_i}}}{t_j}t_i^{ - 1}} \right)} \right|\right)$. A new KF is chosen when $ w_f f + w_{f_t} f_t + w_a a > T_{\text{kf}}$. Here the symbols have the same meaning as in \cite{DSO}.

\textbf{Candidate point tracking:} Candidate points are selected using the approach described in Sec.\ref{sec:points}. These points are then tracked using discrete search along epipolar line and minimizing the photometric error $E_{\text{photo}}$ given by Eq. \ref{eq:error}. The computed depth and co-variance is used to constrain the search interval for the subsequent frame as described in \cite{LSD_SLAM}.

\textbf{Outlier rejection and occlusion detection:} Point observations which have a $E_{\text{photo}}$ above a certain threshold are removed as outliers and excluded for further computation.

\textbf{Parameters initialization:} This step provides the initial estimates of all parameters for optimizing the non convex error $E_{\text{photo}}$. The initial camera pose is computed from direct image alignment and the initial point's depth is from candidate point tracking.

\textbf{Candidate point activation:} New candidates points replace the old marginalized points. The new points are chosen by projecting onto the current frame and maximizing the distance between projection of any existing active points.

\textbf{Marginalization:} This step decides which points and frames should be marginalized. A KF will be marginalized if less than $5\%$ of points are visible in the latest frame. If there are more than $N_{f}$ (fixed at $7$) KFs, a KF which is far from current frame and close to any other KFs will be marginalized.\\

\subsubsection{Back-end}
The back end contains a factor graph which performs continuous windowed optimization using the approach given in \cite{windowed_optimization}. It optimizes $E_{\text{photo}}$ using Gaussian-Newton algorithm in a sliding window manner. The error functions are defined as the following:

For a single active point $p$, its photometric error on KF $j$ is defined as:
\begin{equation}
    E_{pj} = \sum_{p \in N_{p}} w_{p}\left \| (I_{j}[p'] - b_{j}) - \frac{t_{j}e^{a_{j}}}{t_{i}e^{a_{i}}}(I_{i}([p] - b_{i}) \right \|_{\gamma}
\end{equation}
where $p'$ is the projection of point $p$ on KF $j$, $\left\{t_{i}, t_{j}\right\}$ are the exposure time for images $\left\{I_{i}, I_{j}\right\}$, $\left \|  \right \|_\gamma$ is the Huber norm, $a_{i}, a_{j}, b_{i}, b_{j}$ are brightness transfer function parameters, $N_{p}$ is the residual pattern with eight surrounding neighbors and gradient depending weights $w_{p}$ is given by
\begin{equation}
    w_{p} = \frac{c^{2}}{c^{2} + \left \| \bigtriangledown I_{i}(p) \right \|_{2}^{2}}
\end{equation}

The full photometric error over all active points and KFs is defined as
\begin{equation}
    E_{\text{photo}} = \sum_{i \in F} \sum_{p \in P_{i}} \sum_{j \in obs(p)} E_{pj}
    \label{eq:error}
\end{equation}
where $F$ indicates all active KFs, $P_{i}$ indicates all active points in KF $i$, $obs(p)$ indicates all frames' observation in which point $p$ is visible.\\

\section{Point selection based on visual saliency and scene parsing}
\label{sec:PointSel}
\subsection{Visual Saliency Prediction}
Visual saliency is defined as the amount of attention a human would give to each pixel in an image. This is quantitatively measured as the average time a person's gaze rests on each pixel in the image. Prediction of saliency is a hard problem and data driven approaches have lately excelled at this task. We adopt SalGAN \cite{SalGAN} for saliency prediction in SalientDSO. In brief, SalGAN introduced the use of Generative Adversarial Network (GAN)\cite{GAN} for saliency prediction. It contains a generator and a discriminator. The generator is a deep CNN trained on adversarial loss ($L_{GAN}$ in Eq. \ref{eq:adv_loss}), which includes Binary Cross-Entropy loss ($L_{BCE}$ in Eq. \ref{eq:BCE}) to produce a down-sampled saliency map, and the discriminator is a shallower network as compared to the generator, this is trained to solve binary classification between saliency map produced by generator and the groundtruth. The generator is an encoder-decoder type of network, in which the encoder part is identical to VGG-16\cite{VGG16} and its weights are initialized with the weights trained on the ImageNet dataset\cite{ImageNet}. The discriminator's weights are randomly initialized. The whole network is trained on the SALICON dataset\cite{SALICON}.

The adversarial loss is defined as:
\begin{equation}
    L_{GAN} = \alpha \cdot L_{BCE} - logD(I, \hat{S})
    \label{eq:adv_loss}
\end{equation}
where $D(I, \hat{S})$ is the probability of fooling the discriminator. Also, the binary cross-entropy loss between the predicted saliency map $\hat{S}$ and the ground-truth $S$ is defined as:
\begin{equation}
    L_{BCE} = - \frac{1}{N} \sum_{j = 1}^{N} S_{j}log(\hat{S_{j}}) + (1 - S_{j})log(1 - \hat{S_{j}})
    \label{eq:BCE}
\end{equation}
where $S_{j}$ is the probability of pixel $I_{j}$ being fixated.

\begin{figure}[t!]
    \centering
    \includegraphics[width=\columnwidth]{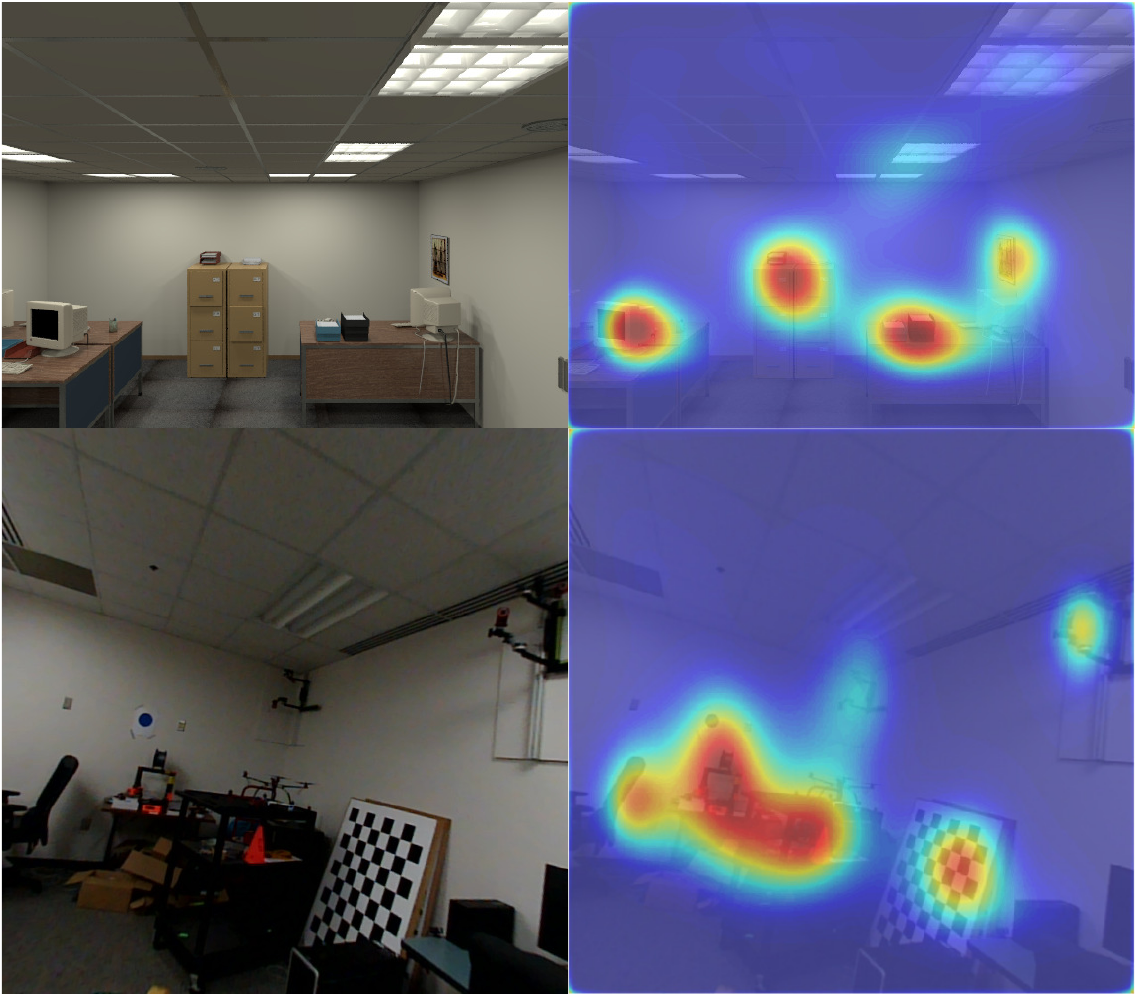}
    \caption{Left column: Input image, Right column: Saliency overlayed on input image.}
    \label{fig:Saliency}
\end{figure}

Some sample results are shown in Fig. \ref{fig:Saliency}. One can clearly notice that walls, floors, and ceilings have lower probability of being fixated on, which is the main idea of the proposed framework.

\subsection{Filtering saliency using semantic information}
The saliency produced by SalGAN is concentrated around a fixation point inside the object and is fuzzy. Moreover, the saliency map is not very robust to viewpoint and illumination changes as the fixation point does not remain constant. In this subsection, we utilize semantic information to filter the saliency. The idea is to weigh down the saliency of uninformative regions, such as walls, ceilings and floors.

To obtain semantic information from a scene, we adopt Pyramid Scene Parsing \cite{PSP} for retrieving semantic labels of every pixel in an image. In brief,  Pyramid Scene Parsing (PSPNet) is a deep neural network for pixel-level prediction tasks. PSPNet uses CNN layers to extract features, then a pyramid parsing module is applied to harvest different sub-region representation, followed by up-sampling and concatenation layers to form the final feature representation. The final features are then fed into more CNN layers to obtain a pixel-level prediction.

Once the per-pixel semantic information $C$ is obtained, the predicted saliency map $\hat{S}$ is filtered by:
\begin{equation}
    \hat{S}^{\text{weighted}}_{j} = w_{C}(C_{j})\hat{S}_{j}
\end{equation}

Here, $w_{C}$ are the predefined weights obtained empirically for different classes. To smooth and maintain a consistent saliency map for each class, each pixel is replaced by the median of saliency for its respective class:
\begin{equation}
    \hat{S}^{\text{final}}_{j} = \text{median}\left \{ \hat{S}^{\text{weighted}}_{i}, \forall i \in C_{j} \right \}
\end{equation}
All steps to generate $\hat{S}^{\text{final}}$ are summarized in Algorithm \ref{alg:saliency}.

\begin{algorithm}[t!]
\KwData{Input image $I$, Pre-defined weights $w_{C}$}
\KwResult{Predicted final saliency $\hat{S}^{\text{final}}$}
$\hat{S}$ = SalGAN($I$)\;
$C$ = PSPNet($I$)\;
\For{ $\forall \left\{ x_j, y_j\right\} \in I$} {
    $\hat{S}^{\text{weighted}}_{j} = w_{C}(C_{j})\hat{S}_{j}$\;
}
\For{$\forall \left\{ x_j, y_j\right\} \in I$} {
    $\hat{S}^{\text{final}}_{j} = \text{median}\left \{ \hat{S}^{\text{weighted}}_{i}, \forall i \in C_{j} \right \}$\;
}
 \caption{Saliency prediction and filtering.}
 \label{alg:saliency}
\end{algorithm}

\begin{algorithm}[t!]
\KwData{Desired number of points $N_{\text{des}}, s_{\text{smooth}}, \hat{S}^{\text{final}}$}
\KwResult{Selected points}
Initialize selected point set as $\left\{ \emptyset\right\}$, $N_{\text{sel}} = 0$\;
 \While{$N_{\text{sel}} < N_{\text{des}}$}{
  Randomly select a patch $M$ from distribution $P_{S}$\;
  Split $M$ into $d \times d$ blocks\;
  \For{each $4d \times 4d$ block}{
    \For{each $2d \times 2d$ block}{
        \For{each $d \times d$ block}{
            Select a point with the highest gradient which surpass the gradient threshold\;
        }
        \If{no selected point in this block}{
            Select a point with the highest gradient which surpass the weaker gradient threshold\;
        }
    }
    \If{no selected point in this block}{
            Select a point with the highest gradient which surpass the much weaker gradient threshold\;
        }
  }
  $N_{\text{sel}} = N_{\text{sel}}\ +$ the number of selected points\;
 }
 \caption{Saliency based points selection.}
 \label{alg:selection}
\end{algorithm}

\subsection{Features/Points selection}
\label{sec:points}
Instead of uniformly selecting candidate points from an image as in DSO, we select points based on saliency. This is very helpful where the scene has a lot of objects or clutter which can be found generally in indoor scenes.

First, we split an image into $K \times K$ patches. For a patch $M_{i}$, we not only compute the median of gradient as a region-adaptive threshold, but also compute the median of saliency as a region-adaptive sampling weight $sw_{i}$. Therefore, for each patch, the sampling weight $sw_{i}$ is computed as:
\begin{equation}
    sw_{i} = \text{median}\left \{ \hat{S}^{\text{final}}_{j}, \forall j \in M_{i} \right \} + s_{\text{smooth}}
\end{equation}
where $s_{\text{smooth}}$ is a laplacian smoothing term used to control the bias on a salient region and the probability of a patch $M_{i}$ being sampled is:
\begin{equation}
    \boldsymbol{P}_{S}(M_{i}) = \frac{sw_{i}}{\sum_{m \in M} sw_{m}}
\end{equation}

Secondly, once a patch $M_{i}$ has been selected, we further split $M_{i}$ into $d \times d$ blocks. For each block, we select the pixel with the highest gradient only if it surpasses the region-adaptive threshold. With this strategy, we can select points which are well distributed in this salient region. In order to extract information from where no high-gradient pixels are present, we follow the same approach as DSO and run two more passes to select pixels with weaker gradient in a larger sub-region with a lower gradient threshold and an increased $d$. A summary of the whole selection method is given in Algorithm \ref{alg:selection}.

\begin{figure}[t!]
    \centering
    \includegraphics[width=\columnwidth]{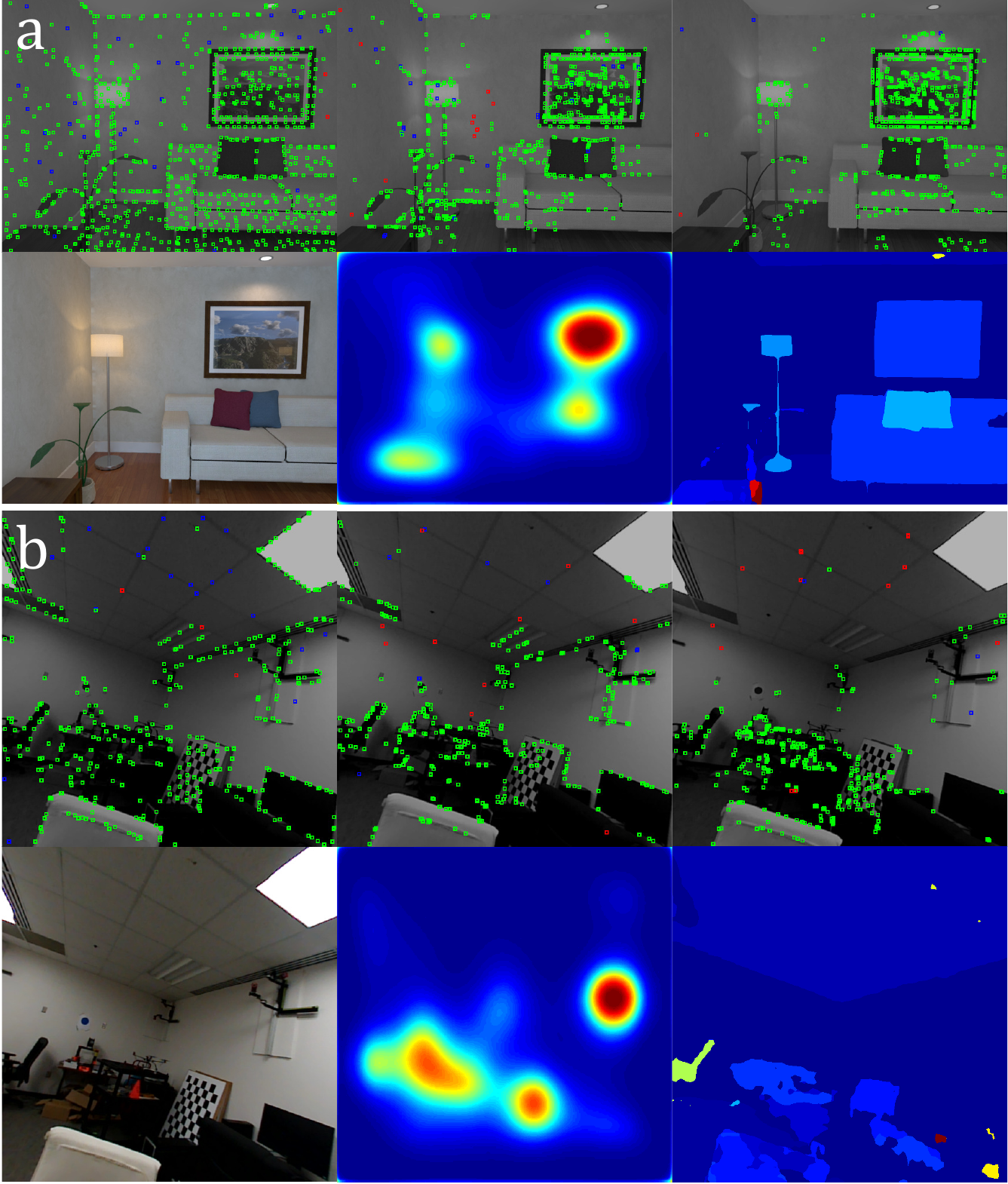}
    \caption{Point selection using different schemes. Top rows in (a) and (b), left to right: features selected using DSO's scheme, saliency only, saliency+scene parsing. Bottom rows in (a) and (b), left to right: input image, saliency, scene parsing output. Notice how using saliency+scene parsing removed all non-informative features.}
    \label{fig:ICLL0Selection}
\end{figure}

Fig. \ref{fig:ICLL0Selection} shows the selected points for some example scenes. We compare our selection based on saliency to the uniform selection adopted by DSO. One can easily notice that texture-less and mostly identical parts, such as walls, floors and ceilings, are down weighted in our pipeline. As demonstrated in Section \ref{sec:results}, this helps us trade the weak features on the floors and ceilings for weak features on objects where the saliency is generally higher - thus, in-turn, making the feature selection more robust and object-centric.

\section{Experimental Results and Analysis}
\label{sec:results}
In this section, we comprehensively evaluate SalientDSO on various datasets.


\begin{itemize}
    \item \textbf{ICL-NUIM dataset}\cite{ICL_NUIM}: This dataset provides two scenes and four different trajectories for each scene which are obtained by running Kintinuous on real image data and finally used in a synthetic framework for obtaining ground-truth.
    \item \textbf{TUM monoVO dataset}\cite{TUM}: This dataset provides 50 sequences comprising over 100 minutes videos. It ranges from indoor corridors to wide outdoor scenes. In our experiments, we only evaluate all methods on indoor sequences $\left\{\text{sequence\_}(1 - 18, 26, 28, 35 - 38, 40)\right\}$. Only the indoor sequences are chosen because the usage of saliency obtained by human gaze is meaningful only for indoor cluttered scenes.
    \item \textbf{CVL dataset}: This dataset was collected by the authors of this paper is available at \url{prg.cs.umd.edu/SalientDSO.html}. The data was collected using a Parrot$^\text{\textregistered}$ SLAMDunk \cite{SLAMDunk} sensor suite. The data from the left camera is used in the experiments.
\end{itemize}

Different parameters used for running the experiments are shown in Table. \ref{table:Parameters}. For ICL-NUIM dataset, photometric correction is not required. To comprehensively evaluate the proposed method, we run each sequence in both forward and backward direction 10 times.

\begin{table}[t!]
\centering
\caption{Parameter settings for different datasets.}
\label{table:Parameters}
\begin{tabular}{llll}
\hline
                                   & TUM  & ICL-NUIM & CVL     \\ \hline
Num of active keyframes $N_{f}$   & 7    & 7  & 7          \\
Num of active points $N_{p}$      & 2000 & 2000 & 1200         \\
Global gradient constant $g_{th}$ & 7    & 3  & 7          \\
Patch size $K$ & 8    & 8    & 8        \\
Photometric correction             & Yes  & Not required & Not available \\ \hline
\end{tabular}
\end{table}

\subsection{Quantitative Evaluation}
\begin{figure}[t!]
    \centering
    \includegraphics[width=\columnwidth]{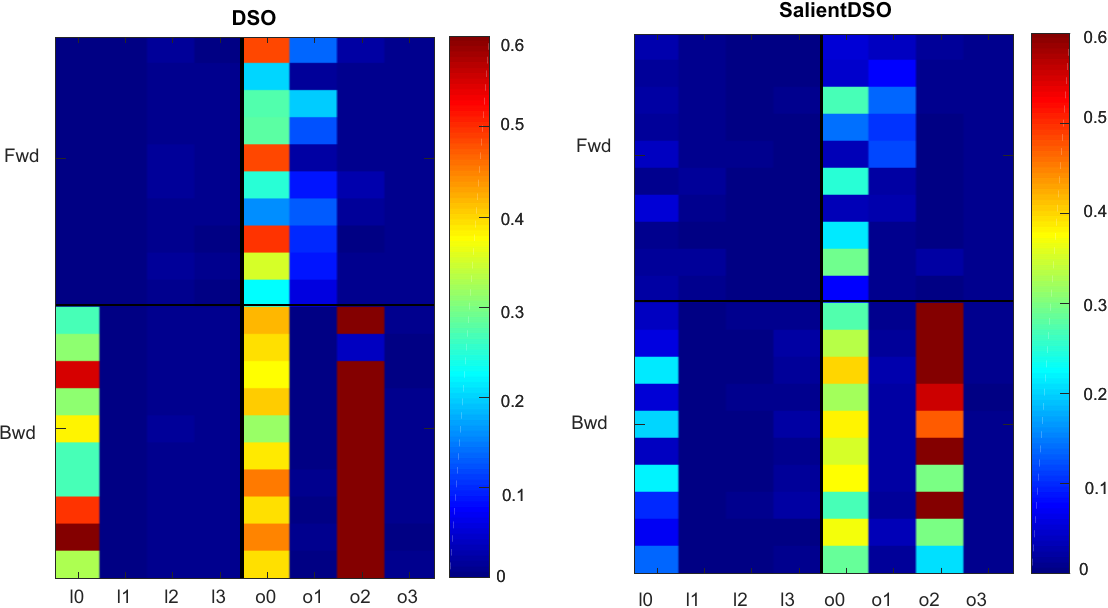}
    \caption{Comparison of evaluation results for ICL-NIUM dataset. Left: DSO, Right: SalientDSO. Each square correspondes to a color coded error. Note that Salient DSO almost always has lower error than it's DSO counterpart.}
    \label{fig:ICL_err}
\end{figure}

\begin{figure}[t!]
    \centering
    \includegraphics[width=\columnwidth]{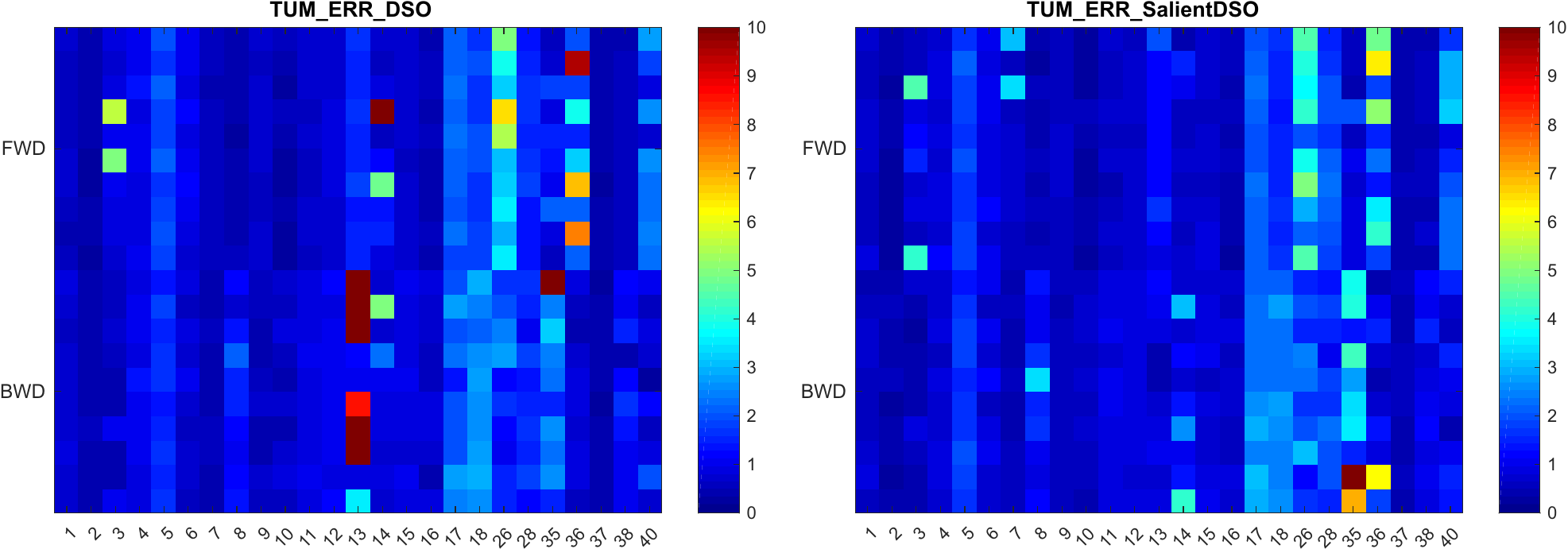}
    \caption{Comparison of evaluation results for TUM dataset. Left: DSO, Right: SalientDSO. Note that Salient DSO almost always has lower error than it's DSO counterpart. Note that, for the TUM dataset scene parsing was turned off as TUM dataset only provoides grayscale images and scene parsing outputs are very noisy for grayscale images.}
    \label{fig:TUM_err}
\end{figure}

Fig. \ref{fig:ICL_err} shows the absolute trajectory Root Mean Square Error (RMSE$_{\text{ate}}$) on ICL-NUIM dataset. Using visual saliency driven features, SalientDSO performs better in accuracy as compared to DSO. We also report alignment error $e_{\text{align}}$ on TUM monoVO dataset in Fig. \ref{fig:TUM_err}. We disable the semantic filtering when we evaluate the proposed method on the TUM monoVO dataset, since this dataset provides only grayscale images and outputs from PSPNet are inaccurate and noisy for grayscale images. In Tables \ref{table:ErrICL} and \ref{table:ErrTUM}, we compare our method to DSO and ORB-SLAM on the ICL-NUIM and TUM monoVO datasets. DSO and ORB-SLAM are the current state-of-the-art direct and feature-based monocular VO methods. The results for DSO and ORB-SLAM are taken from \cite{DSO}. ORB-SLAM is a full-fledged SLAM framework with loop closure and global alignment, while DSO and SalientDSO are merely odometry frameworks. To make the comparison fair, loop-closure detection and re-localization have been turned off for ORM-SLAM. The missing values in the table represent tracking failures. We achieve similar or better performance on most sequences. The improvement is not significant on the TUM monoVO dataset because most of the sequences involve a traversal through a hallway where there are no local salient objects or features for saliency prediction to work well. This makes SalientDSO's performance close to that of traditional DSO.  

\begin{table}[]
\centering
\caption{RMSE$_{\text{ate}}$ on ICL-NIUM dataset in m.}
\label{table:ErrICL}
\resizebox{\columnwidth}{!}{
\begin{tabular}{lllllll}
\toprule
        & \multicolumn{3}{c}{Forward} & \multicolumn{3}{c}{Backward}     \\
   Sequence     & ORB      & DSO & SalientDSO & ORB      & DSO & SalientDSO      \\ \hline
ICL\_l0 & 0.01     & \textbf{0.003} & 0.022    & \textbf{0.01}     & -     & 0.112         \\
ICL\_l1 & 0.02     & \textbf{0.004} & 0.009    & 0.04     & 0.003 & \textbf{0.003}         \\
ICL\_l2 & 0.06     & 0.012 & \textbf{0.004}    & 0.19     & 0.010 & \textbf{0.005}         \\
ICL\_l3 & 0.03     & 0.006 & \textbf{0.004}    & 0.05     & \textbf{0.008} & 0.013         \\
ICL\_o0 & 0.21     & 0.320 & \textbf{0.140}    & 0.41     & 0.399 & \textbf{0.336}         \\
ICL\_o1 & 0.83     & 0.094 & \textbf{0.055}    & 0.68     & \textbf{0.006} & 0.020         \\
ICL\_o2 & 0.37     & 0.012 & \textbf{0.008}    & 0.32     & 0.582 & \textbf{0.512}         \\
ICL\_o3 & 0.65     & \textbf{0.007} & 0.009    & 0.06     & \textbf{0.006} & 0.008         \\
\hline
\textbf{Overall Avg.} & 0.271    & 0.057 & \textbf{0.031}   & 0.218    & 0.144$^*$     &
\textbf{0.126}        \\ 
\bottomrule
\end{tabular}}
  \\\scriptsize{$^*$ indicates average taken only on sequences which completed.}
\end{table}

\begin{table}[]
\centering
\caption{$e_{\text{align}}$ on TUM monoVO dataset in m.}
\label{table:ErrTUM}
\resizebox{\columnwidth}{!}{
\begin{tabular}{lllllll}
\toprule
        & \multicolumn{3}{c}{Forward} & \multicolumn{3}{c}{Backward}     \\
     Sequence   & ORB      & DSO & SalientDSO & ORB      & DSO & SalientDSO      \\ \hline
seq\_01 & 3.02     & \textbf{0.59} & 0.60      & 1.73     & 0.72 & \textbf{0.60}           \\
seq\_02 & 16.12    & 0.36 & \textbf{0.33}      & 3.23     & \textbf{0.43} & 0.44           \\
seq\_03 & 3.42     & 1.75 & \textbf{1.55}      & 1.42     & 0.59 & \textbf{0.50}           \\
seq\_04 & 9.95     & 0.98 & \textbf{0.82}      & 5.95     & 1.00 & \textbf{0.76}           \\
seq\_05 & -        & 1.86 & \textbf{1.77}      & -        & \textbf{1.55} & 1.66           \\
seq\_06 & -        & 0.97 & \textbf{0.93}      & 1.25     & \textbf{0.73} & 0.81           \\
seq\_07 & 1.69     & \textbf{0.55} & 1.14      & 2.02     & \textbf{0.44} & 0.48           \\
seq\_08 & 436.00   & \textbf{0.36} & 0.44      & 2.63     & \textbf{1.28} & 1.47           \\
seq\_09 & 2.04     & 0.65 & \textbf{0.58}      & 0.67     & \textbf{0.52} & 0.53           \\
seq\_10 & 2.52     & 0.35 & \textbf{0.34}      & 1.43     & 0.61 & \textbf{0.61}           \\
seq\_11 & 7.20     & 0.62 & \textbf{0.58}      & 2.99     & \textbf{0.87} & 0.89           \\
seq\_12 & 2.98     & 0.75 & \textbf{0.67}      & 3.10     & 1.01 & \textbf{0.84}           \\
seq\_13 & 5.13     & 1.54 & \textbf{1.27}      & 2.59     & 8.96 & \textbf{0.81}           \\
seq\_14 & 13.27    & 2.89 & \textbf{0.71}      & 2.10     & \textbf{1.35} & 1.69           \\
seq\_15 & 2.90     & 0.71 & \textbf{0.71}      & 1.90     & 0.88 & \textbf{0.81}           \\
seq\_16 & 2.40     & 0.47 & \textbf{0.45}      & 1.58     & 0.72 & \textbf{0.67}           \\
seq\_17 & 12.29    & 2.10 & \textbf{2.10}      & \textbf{1.50}     & 2.13 & 2.50           \\
seq\_18 & 14.64    & 1.77 & \textbf{1.52}      & -        & 2.62 & \textbf{2.47}           \\
seq\_26 & 28.46    & 3.98 & \textbf{3.60}      & 4.62     & \textbf{1.66} & 1.89           \\
seq\_28 & 19.17    & \textbf{1.48} & 1.88      & 3.57     & \textbf{1.47} & 1.65           \\
seq\_35 & 14.09    & 1.10 & \textbf{0.84}      & 16.81    & \textbf{5.48} & 9.97           \\
seq\_36 & 1.81     & 4.01 & \textbf{3.25}      & 1.69     & \textbf{0.70} & 1.46           \\
seq\_37 & 0.60     & \textbf{0.35} & 0.40      & 1.30     & \textbf{0.37} & 0.46           \\
seq\_38 & -        & 0.55 & \textbf{0.50}      & 24.77    & 1.10 & \textbf{1.03}           \\
seq\_40 & -        & \textbf{2.04} & 2.16      & 18.93    & \textbf{0.87} & 1.04           \\
\hline 
\textbf{Overall Avg.} & 28.55$^*$        & 1.31 & \textbf{1.17}      & -        & 1.52 & \textbf{1.44} \\
\bottomrule
\end{tabular}}
  \\\scriptsize{$^*$ indicates average taken only on sequences which completed.}
\end{table}

\begin{figure}[t!]
    \centering
    \includegraphics[width=\columnwidth]{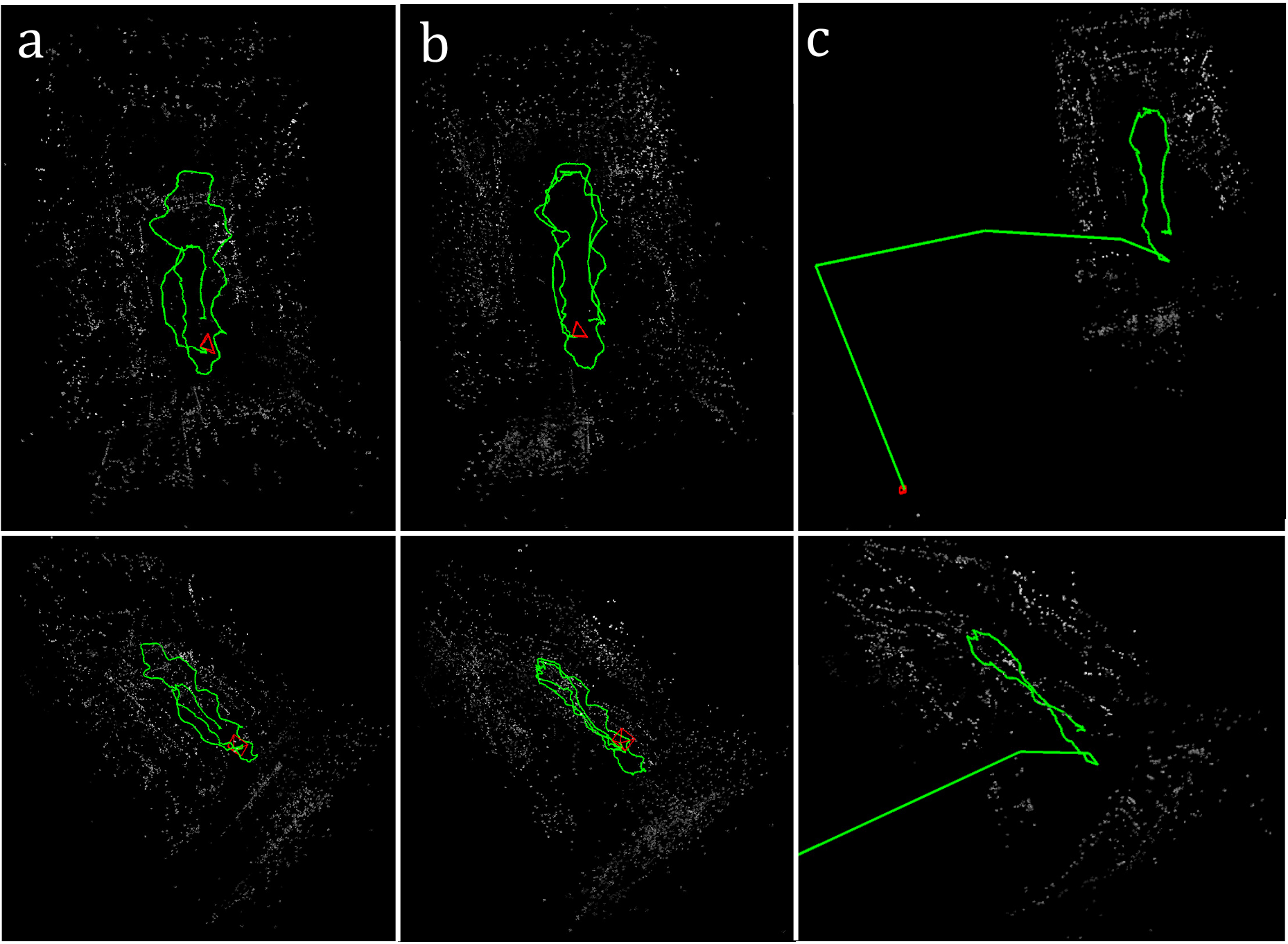}
    \caption{Comparison of outputs for $N_p=40$ -- very few features. (a) Success case of DSO with a large amount of drift, (b) Success case for SalientDSO,  (c) Failure case of DSO where the optimization diverges due to very few features. Notice that SalientDSO can perform very well in these extreme conditions showing the robustness of the features chosen.}
    \label{fig:CVL_N30}
\end{figure}

The claim in the paper is that the usage of visual saliency should result in more robust features than just using image gradient based features as in DSO. The intuition behind this claim is that visual saliency includes high level semantics which inherently make the features more robust. To support this claim, we anticipate that SalientDSO should perform much better than DSO when the number of points is very low (as low as 40 points). To demonstarate this claim, we evaluate on each CVL sequence. We run each sequence in both forward and backward direction 100 times, with an extremely low point density of $N_{p} = 40$. The results are shown in Table. \ref{table:CVL}. We define failure as either an optimization failure or tracking loss. Our proposed method is much more robust and predicts an accurate trajectory, while DSO has a much higher failure rate and its trajectory and projected point cloud shows significant drift in scale and position. An example of trajectory and projected point cloud is shown in Fig. \ref{fig:CVL_N30}. This experiment highlights the robustness of features chosen in SalientDSO for cluttered indoor scenes and how this will be useful for robots with very low computation power due to the less computational and memory requirements when $N_p$ is low.

\begin{table}[t!]
\centering
\caption{Comparison of success rate between DSO and SalientDSO on CVL dataset.}
\label{table:CVL}
\begin{tabular}{lll}
\hline
 Sequence   & DSO & SalientDSO \\ \hline
CVL\_01\_Fwd & 53\%  & \textbf{65\%}         \\
CVL\_01\_Bwd & 59\%  & \textbf{92\%}         \\
CVL\_02\_Fwd & 73\%  & \textbf{96\%}         \\
CVL\_02\_Bwd & 71\%  & \textbf{91\%}         \\\hline
\end{tabular}
\end{table}

\subsection{Qualitative Evaluation}
\begin{figure}[t!]
    \centering
    \includegraphics[width=\columnwidth]{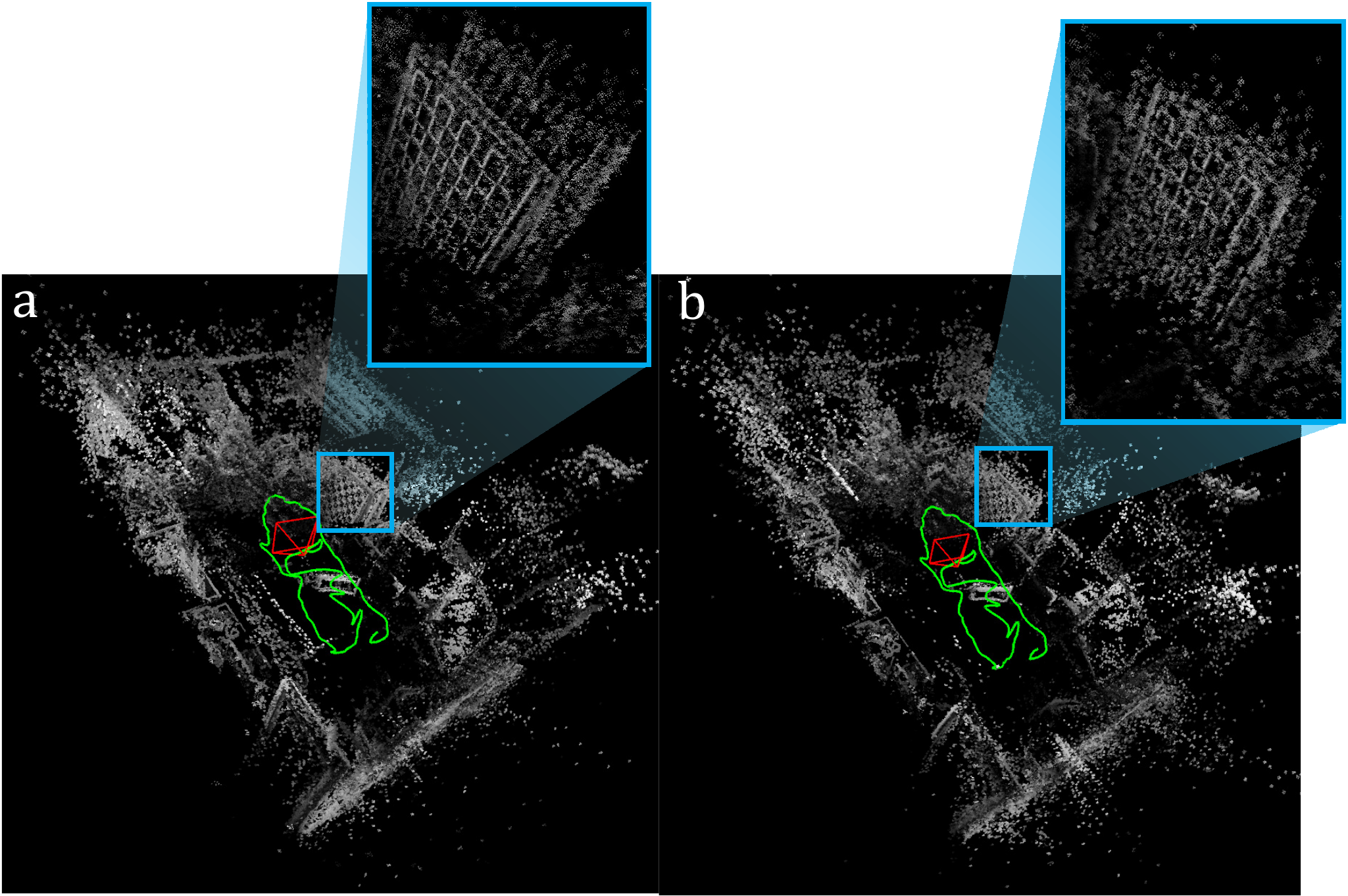}
    \caption{Comparison of drift. (a) DSO, (b) SalientDSO. Observe that SalientDSO's output has the checkerboard from different times more closely aligned as compared to DSO. Here $N_p=1000$.}
    \label{fig:CVL1000Pts}
\end{figure}

\begin{figure}[t!]
    \centering
    \includegraphics[width=\columnwidth]{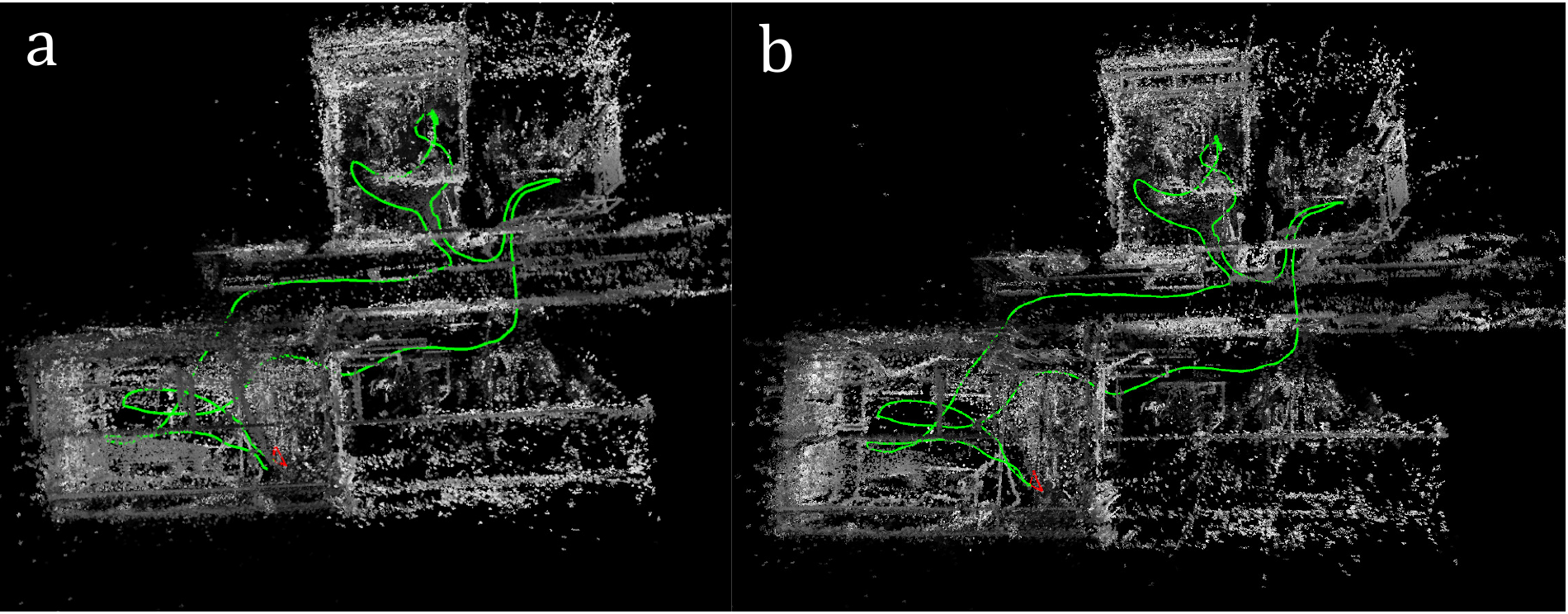}
    \caption{Sample outputs for TUM sequence\_1. (a) DSO, (b) SalientDSO. Here $N_p=1000$.}
    \label{fig:TUM}
\end{figure}

Examples of the reconstructed scenes of sequences CVL\_01 and TUM $\text{seqence}\_01$ are shown in Figs. \ref{fig:CVL1000Pts} and \ref{fig:TUM} respectively. Although both reconstructed scenes look similar, one could observe that amount of drift in SalientDSO is much less compared to DSO (refer to the zoomed part of  Fig. \ref{fig:CVL1000Pts}). One can clearly observe that the checkerboard of different loops align better in our approach. Instead of sampling random high gradient points, sampling salient and important points improves the robustness of VO. Sampling salient points achieves removing outliers and points with unconstrained depth in optimization which improves the prediction of initial estimates and the output of windowed bundle adjustment in optimization.

\section{Conclusions}
\label{sec:Conc}
We introduce the philosophy of attention and fixation to visual odometry. Based on this philosophy, we develop Salient Direct Sparse Odometry, which brings the concept of attention and fixation based on visual saliency into Visual Odometry to achieve robust feature selection. We provide thorough quantitative and qualitative evaluations on ICL-NUIM and TUM monoVO dataset to demonstrate that using salient features improves the robustness and accuracy. We also collect and publicly release a new CVL dataset with cluttered scenes for mapping. We show the robustness of our features by very low drift visual odometry with as low as 40 features per frame. Our method takes about a second per frame for computation of saliency and scene parsing on an NVIDIA Titan-Xp GPU and the remaining computations run real-time at 30fps on an Intel$^\text{\textregistered}$ Core i7 6850K 3.6GHz CPU. In the near future, we plan to extend our method to outdoor environment. We also consider to implement our method on hardware to make the complete pipeline real-time.

\section*{Acknowledgement}
This work was partly funded by the Brin Family Foundation, Northrop Grumman Corporation and National Science Foundation under grant SMA 1540917 and grant CNS 1544797. 

\bibliographystyle{unsrt}
\bibliography{Ref}

\begin{thebibliography}{10}

\bibitem{monoslam}
Andrew~J Davison, Ian~D Reid, Nicholas~D Molton, and Olivier Stasse.
\newblock Monoslam: Real-time single camera slam.
\newblock {\em IEEE transactions on pattern analysis and machine intelligence},
  29(6):1052--1067, 2007.

\bibitem{PTAM}
Georg Klein and David Murray.
\newblock Parallel tracking and mapping for small ar workspaces.
\newblock In {\em Mixed and Augmented Reality, 2007. ISMAR 2007. 6th IEEE and
  ACM International Symposium on}, pages 225--234. IEEE, 2007.

\bibitem{DTAM}
Richard~A Newcombe, Steven~J Lovegrove, and Andrew~J Davison.
\newblock Dtam: Dense tracking and mapping in real-time.
\newblock In {\em Computer Vision (ICCV), 2011 IEEE International Conference
  on}, pages 2320--2327. IEEE, 2011.

\bibitem{LSDSLAM}
Jakob Engel, Thomas Sch{\"o}ps, and Daniel Cremers.
\newblock Lsd-slam: Large-scale direct monocular slam.
\newblock In {\em European Conference on Computer Vision}, pages 834--849.
  Springer, 2014.

\bibitem{ORBSLAM}
R.~Mur-Artal, J.~M.~M. Montiel, and J.~D. Tardós.
\newblock Orb-slam: A versatile and accurate monocular slam system.
\newblock {\em IEEE Transactions on Robotics}, 31(5):1147--1163, Oct 2015.

\bibitem{stuhmer2010real}
Jan St{\"u}hmer, Stefan Gumhold, and Daniel Cremers.
\newblock Real-time dense geometry from a handheld camera.
\newblock In {\em Joint Pattern Recognition Symposium}, pages 11--20. Springer,
  2010.

\bibitem{ROVIO}
Michael Bloesch, Sammy Omari, Marco Hutter, and Roland Siegwart.
\newblock Robust visual inertial odometry using a direct ekf-based approach.
\newblock In {\em Intelligent Robots and Systems (IROS), 2015 IEEE/RSJ
  International Conference on}, pages 298--304. IEEE, 2015.

\bibitem{RGBDSLAM}
Sebastian~A Scherer and Andreas Zell.
\newblock Efficient onbard rgbd-slam for autonomous mavs.
\newblock In {\em Intelligent Robots and Systems (IROS), 2013 IEEE/RSJ
  International Conference on}, pages 1062--1068. IEEE, 2013.

\bibitem{DSO}
J.~Engel, V.~Koltun, and D.~Cremers.
\newblock Direct sparse odometry.
\newblock {\em IEEE Transactions on Pattern Analysis and Machine Intelligence},
  Apr 2017.

\bibitem{YiannisFixation}
Ajay Mishra, Yiannis Aloimonos, and Cheong~Loong Fah.
\newblock Active segmentation with fixation.
\newblock In {\em Computer Vision, 2009 IEEE 12th International Conference on},
  pages 468--475. IEEE, 2009.

\bibitem{ActiveVision}
John Aloimonos et~al.
\newblock Active vision.
\newblock {\em International journal of computer vision}, 1(4):333--356, 1988.

\bibitem{SukhtameActive}
Jeannette Bohg et~al.
\newblock Interactive perception: Leveraging action in perception and
  perception in action.
\newblock {\em IEEE Transactions on Robotics}, 33(6):1273--1291, 2017.

\bibitem{BajcsyActive}
Ruzena Bajcsy et~al.
\newblock Revisiting active perception.
\newblock {\em Autonomous Robots}, pages 1--20, 2017.

\bibitem{SemanticSLAM}
Javier Civera, Dorian G{\'a}lvez-L{\'o}pez, Luis Riazuelo, Juan~D Tard{\'o}s,
  and JMM Montiel.
\newblock Towards semantic slam using a monocular camera.
\newblock In {\em Intelligent Robots and Systems (IROS), 2011 IEEE/RSJ
  International Conference on}, pages 1277--1284. IEEE, 2011.

\bibitem{KostasSemanticSLAM}
Sean~L Bowman, Nikolay Atanasov, Kostas Daniilidis, and George~J Pappas.
\newblock Probabilistic data association for semantic slam.
\newblock In {\em Robotics and Automation (ICRA), 2017 IEEE International
  Conference on}, pages 1722--1729. IEEE, 2017.

\bibitem{an2017semantic}
Lifeng An, Xinyu Zhang, Hongbo Gao, and Yuchao Liu.
\newblock Semantic segmentation--aided visual odometry for urban autonomous
  driving.
\newblock {\em International Journal of Advanced Robotic Systems},
  14(5):1729881417735667, 2017.

\bibitem{alexis}
Kostas~Alexis Tung~Dang, Christos~Papachristos.
\newblock Visual saliency–aware receding horizon autonomous exploration with
  application to aerial robotics.
\newblock In {\em Robotics and Automation (ICRA), 2018 IEEE International
  Conference on}. IEEE, 2018.

\bibitem{LSD_SLAM}
J.~Engel, T.~Sch\"ops, and D.~Cremers.
\newblock {LSD-SLAM}: Large-scale direct monocular {SLAM}.
\newblock In {\em European Conference on Computer Vision (ECCV)}, September
  2014.

\bibitem{windowed_optimization}
Stefan Leutenegger, Simon Lynen, Michael Bosse, Roland Siegwart, and Paul
  Furgale.
\newblock Keyframe-based visual–inertial odometry using nonlinear
  optimization.
\newblock {\em The International Journal of Robotics Research}, 34(3):314--334,
  2015.

\bibitem{SalGAN}
J.~{Pan}, C.~{Canton Ferrer}, K.~{McGuinness}, N.~E. {O'Connor}, J.~{Torres},
  E.~{Sayrol}, and X.~{Giro-i-Nieto}.
\newblock {SalGAN: Visual Saliency Prediction with Generative Adversarial
  Networks}.
\newblock {\em ArXiv e-prints}, January 2017.

\bibitem{GAN}
Ian Goodfellow, Jean Pouget-Abadie, Mehdi Mirza, Bing Xu, David Warde-Farley,
  Sherjil Ozair, Aaron Courville, and Yoshua Bengio.
\newblock Generative adversarial nets.
\newblock In Z.~Ghahramani, M.~Welling, C.~Cortes, N.~D. Lawrence, and K.~Q.
  Weinberger, editors, {\em Advances in Neural Information Processing Systems
  27}, pages 2672--2680. Curran Associates, Inc., 2014.

\bibitem{VGG16}
K.~{Simonyan} and A.~{Zisserman}.
\newblock {Very Deep Convolutional Networks for Large-Scale Image Recognition}.
\newblock {\em ArXiv e-prints}, September 2014.

\bibitem{ImageNet}
J.~Deng, W.~Dong, R.~Socher, L.-J. Li, K.~Li, and L.~Fei-Fei.
\newblock {ImageNet: A Large-Scale Hierarchical Image Database}.
\newblock In {\em CVPR09}, 2009.

\bibitem{SALICON}
Ming Jiang, Shengsheng Huang, Juanyong Duan, and Qi~Zhao.
\newblock Salicon: Saliency in context.
\newblock In {\em The IEEE Conference on Computer Vision and Pattern
  Recognition (CVPR)}, June 2015.

\bibitem{PSP}
H.~Zhao, J.~Shi, X.~Qi, X.~Wang, and J.~Jia.
\newblock Pyramid scene parsing network.
\newblock In {\em 2017 IEEE Conference on Computer Vision and Pattern
  Recognition (CVPR)}, volume~00, pages 6230--6239, July 2017.

\bibitem{ICL_NUIM}
A.~Handa, T.~Whelan, J.B. McDonald, and A.J. Davison.
\newblock A benchmark for {RGB-D} visual odometry, {3D} reconstruction and
  {SLAM}.
\newblock In {\em IEEE Intl. Conf. on Robotics and Automation, ICRA}, Hong
  Kong, China, May 2014.

\bibitem{TUM}
J.~{Engel}, V.~{Usenko}, and D.~{Cremers}.
\newblock {A Photometrically Calibrated Benchmark For Monocular Visual
  Odometry}.
\newblock {\em ArXiv e-prints}, July 2016.

\bibitem{SLAMDunk}
{Parrot SLAMDunk}.
\newblock \url{http://developer.parrot.com/docs/slamdunk/}, 2018.

\end{thebibliography}
\end{document}